\documentclass[10pt,twocolumn,letterpaper]{article}

\usepackage{wacv}              

\usepackage{graphicx}
\usepackage{amsmath}
\usepackage{amssymb}
\usepackage{booktabs}
\usepackage[absolute]{textpos}
\usepackage[pagebackref,breaklinks,colorlinks]{hyperref}

\usepackage[capitalize]{cleveref}
\crefname{section}{Sec.}{Secs.}
\Crefname{section}{Section}{Sections}
\Crefname{table}{Table}{Tables}
\crefname{table}{Tab.}{Tabs.}

\usepackage{bm}
\usepackage{comment}

\usepackage{siunitx}
\usepackage{graphicx}

\usepackage{booktabs}
\usepackage{tabularx}
\newcolumntype{L}[1]{>{\raggedright\arraybackslash}m{#1}}
\newcolumntype{C}[1]{>{\centering\arraybackslash}m{#1}}
\newcolumntype{R}[1]{>{\raggedleft\arraybackslash}m{#1}}
\usepackage{multirow}
\usepackage{xcolor}

\usepackage[normalem]{ulem}

\newcommand{\abs}[1]{\left\vert#1\right\vert}
\newcommand{\norm}[1]{\left\Vert#1\right\Vert}

\renewcommand{\vec}[1]{\bm{#1}}
\newcommand{\mat}[1]{\bm{#1}}
\newcommand{\set}[1]{\mathcal{#1}}
\newcommand{\given}{\!\mid\!}

\newcommand{\euler}{\mathrm{e}}

\newcommand{\identity}{\mathrm{I}}

\DeclareMathOperator{\clip}{clip}
\DeclareMathOperator{\lerp}{lerp}
\DeclareMathOperator{\diag}{diag}
\DeclareMathOperator{\sg}{sg}

\let\originalleft\left
\let\originalright\right
\renewcommand{\left}{\mathopen{}\mathclose\bgroup\originalleft}
\renewcommand{\right}{\aftergroup\egroup\originalright}

\def\wacvPaperID{1385} 
\def\confName{WACV}
\def\confYear{2025}

\begin{document}

\title{ROSA: Reconstructing Object Shape and Appearance Textures\\by Adaptive Detail Transfer}

\author{Julian Kaltheuner \qquad Patrick Stotko \qquad Reinhard Klein\\
University of Bonn\\
{\tt\small \{kaltheun,stotko,rk\}@cs.uni-bonn.de }
}
\maketitle
\begin{abstract}
    Reconstructing an object's shape and appearance in terms of a mesh textured by a spatially-varying bidirectional reflectance distribution function (SVBRDF) from a limited set of images captured under collocated light is an ill-posed problem.
    Previous state-of-the-art approaches either aim to reconstruct the appearance directly on the geometry or additionally use texture normals as part of the appearance features.
    However, this requires detailed but inefficiently large meshes, that would have to be simplified in a post-processing step, or suffers from well-known limitations of normal maps such as missing shadows or incorrect silhouettes.
    Another limiting factor is the fixed and typically low resolution of the texture estimation resulting in loss of important surface details.
    To overcome these problems, we present ROSA, an inverse rendering method that directly optimizes mesh geometry with spatially adaptive mesh resolution solely based on the image data.
    In particular, we refine the mesh and locally condition the surface smoothness based on the estimated normal texture and mesh curvature.
    In addition, we enable the reconstruction of fine appearance details in high-resolution textures through a pioneering tile-based method that operates on a single pre-trained decoder network but is not limited by the network output resolution.

\end{abstract}
\begin{textblock}{15}(3,25.8)
{\footnotesize\noindent\color{gray} © 2025 IEEE.  Personal use of this material is permitted.  Permission from IEEE must be obtained for all other uses, in any current or future media, including reprinting/republishing this material for advertising or promotional purposes, creating new collective works, for resale or redistribution to servers or lists, or reuse of any copyrighted component of this work in other works.
}
\end{textblock}
\section{Introduction}
\label{sec:intro}

\begin{figure}[t]
    \centering
    \includegraphics[width=\linewidth]{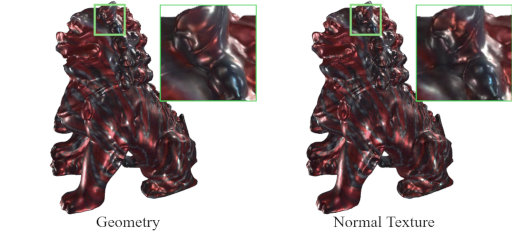}
    \caption{Exemplary renderings in Unity emphasizing artifacts from geometric surface details being approximated in normal textures instead of faithfully representing them in the mesh geometry.}
    \label{fig:comparison}
\end{figure}

The joint recovery of surface geometry and appearance of a real-world object purely from 2D image data is a classic problem in computer vision and computer graphics.
In addition to its great relevance in the context of augmented reality (AR) and virtual reality (VR) such as teleportation scenarios, there exist numerous applications in entertainment such as video games and movies, cultural heritage, and many others.
A particular challenge in ensuring high-quality visualizations lies in the need for accurate geometry as well as detailed material properties in terms of a spatially-varying bidirectional reflectance distribution function (SVBRDF) to allow for relighting.

Neural representations such as neural radiance fields (NeRFs)~\cite{mildenhall2020nerf} have gained significant interest due to their capability of synthesising highly realistic novel views of objects and even complex scenes.
The strong entanglement of appearance and surrounding illumination in its representation, however, prohibits several applications which led to the development of several extensions~\cite{yariv2020multiview,srinivasan2021nerv,boss2021nerd,zhang2021nerfactor}.
Although these approaches address this inherent limitation, they still employ either detailed but very large meshes or implicit volumetric representations that require explicit surface extraction for efficient rendering.
While the shape resolution can effectively be simplified in a post-processing step~\cite{garland1997surface,rossignac1993multi,potamias2022revisiting,potamias2022neural}, the underlying criteria typically only consider errors on a pure geometrical level and do not account for the actually perceived visual quality relative to the originally provided input image data.

A straightforward strategy for obtaining a compact explicit geometry representation without requiring additional simplification would be to directly estimate a lower-resolution mesh while modeling the highly detailed appearance in terms of surface textures.
However, small variations in the appearance that are visible in the input data cannot be faithfully represented in this way since these effects strongly correlate to variations in the surface orientation and, in turn, the geometry.
The high dimensionality of the material properties and the comparably low number of appearance samples per surface point introduces further ambiguities that cannot be handled via direct optimization of all unknown parameters.
In this context, normal maps provide a simple and efficient way to circumvent this problem and can be learned together with SVBRDFs in a low-dimensional latent space~\cite{guo2020materialgan,kaltheuner2021capturing,kaltheuner2023unified}.
While this generally leads to promising results, the resulting reconstructions still suffer from the well-known limitations of normal maps such as wrong silhouettes and missing shadows, see~\cref{fig:comparison}.
Instead, employing the more general displacement mapping technique could resolve this issue, yet optimizing such a map jointly with a low-resolution base mesh is inherently ambiguous and an ill-posed problem as well.

In this paper, we explore strategies for reconstructing objects from a set of images captured under collocated light \emph{entirely} using inverse rendering while addressing the aforementioned problems.
In particular, we represent shape as a parameterized triangle mesh with adaptive resolution and appearance features in form of a texture atlas whose texels contain all material parameters including normals, which are suitable for visualization with standard rendering pipelines.
To avoid local minima in the optimization of the shape, we exploit a normal map which ensures that fine visible details are correctly represented by the textures throughout the whole reconstruction process.
Furthermore, we address the mentioned drawbacks of normal maps by transferring as many geometric details of the captured object as possible from the normal map onto the mesh geometry, in other words, all details of the object that have a size of more than one pixel in the recorded views.
This is realized through an adaptive subdivision of the mesh in areas of high curvature, guided by a novel normal loss term and supplemented by a local smoothness preconditioning term.
For efficient and robust estimation of the appearance information, we decompose the texture atlas into individual tiles whose texture is generated by a single decoder network that is fine-tuned on the tile set during the reconstruction process.
While the tile size is fixed, the decoder can operate on any number of tiles and, thus, even reconstruct high-resolution appearance features.
In particular, we pre-train the decoder as part of an autoencoder network on a large material database to ensure that the fixed, randomly sampled latent codes for each tile correspond to plausible SVBRDFs and can be robustly refined to recover fine texture details.

In summary, the key contributions of our work are:
\begin{itemize}
    \item We present ROSA, an inverse rendering method for reconstructing shape and appearance of objects with adaptive resolution without the need for further post-processing.
    \item We dynamically allocate detail information to either the mesh geometry or the normal map based on detail size using a novel normal loss for optimal balancing.
    \item We obtain appearance textures by finetuning a decoder network and employ a tiling-based approach to allow generating arbitrary-sized textures.
\end{itemize}
The source code of our work is available at \url{https://github.com/vc-bonn/rosa}.

\section{Related Work}
\label{sec:related-work}

\paragraph*{Neural Representations.}
Due to significant advances in differentiable rendering, neural implicit scene representations have gained increased in recent years which enable synthesising realistic novel views~\cite{lombardi2019neural,sitzmann2019scene,niemeyer2020differentiable,bi2020neural,bi2020deep}.
Among these, especially the work on neural radiance fields (NeRFs)~\cite{mildenhall2020nerf} became very popular due to its simple yet powerful scene representation, in terms of a density field as well as a view-dependent radiance field that is encoded by a multi-layer perceptron (MLP), and the generated high-quality visualizations.
Further developments investigated surface representations based on implicit signed distance distance functions (SDF) to replace the density field and improve the accuracy of the geometry~\cite{yariv2020multiview,wang2021neus,vicini2022differentiable,oechsle2021unisurf,yariv2021volume,fan2023factored,liang2023envidr,wu2023nefii,ge2023ref}.
Similarly, more efficient neural representations of the appearance have also been developed.
Based on the seminal work on neural textures~\cite{thies2019deferred}, several approaches employ coarse proxy meshes equipped with a neural texture to enable fast rendering through rasterization pipelines and custom neural shaders required for decoding~\cite{yang2022neumesh,xiang2021neutex,gao2020deferred,chen2023mobilenerf,tang2023delicate}.

We also reconstruct appearance textures in an implicit manner by finetuning a pre-trained decoder network which, however, is utilized to generate realistic SVBRDFs that can directly be rendered by standard graphics engines.

\paragraph*{Shape and SVBRDF Estimation.}
While the aforementioned methods achieve a high degree of realism, the underlying representation of object appearance largely differ from physical models.
Thus, several approaches~\cite{zhang2021nerfactor,boss2021nerd,srinivasan2021nerv,boss2022samurai,jin2023tensoir,sun2023neural,bi2020neural,zhang2022iron} considered a multi-stage approach where an initially trained radiance field is decomposed into a surface shape, surface material parameters in terms of a SVBRDF, as well as surrounding illumination.
Recently, hybrid models defined as volumetric microflake~\cite{zhang2023nemf} and microfacet~\cite{mai2023neural} fields and combined the ray marching procedure of volume rendering with importance-sampled path tracing according to the distribution of the micro structures.
Further methods also considered Monte Carlo-based path tracing~\cite{goel2020shape,luan2021unified,brahimi2024supervol,hasselgren2022shape} to faithfully model light transport during inverse rendering.
To regularize the appearance decomposition, many approaches assume particular common illumination models such as directional lighting~\cite{yang2022psnerf}, environment maps~\cite{srinivasan2021nerv,boss2021nerd,zhang2021nerfactor}, collocated point lights~\cite{bi2020neural,zhang2022iron,brahimi2024supervol,kaltheuner2023unified,li2018learning,boss2020two,luan2021unified,nam2018practical}, or spherical Gaussians~\cite{zhang2021physg,munkberg2022extracting,hasselgren2022shape}.
Controlling the illumination with more general patterns using specifically tailored capturing devices have also been investigated in the past.
This includes movable handheld sensors, where a camera is attached with surrounding point lights~\cite{schmitt2020joint,schmitt2023towards} or an array of small LEDs forming an area light, as well as affordable stationary devices with white environment illumination~\cite{kang2019learning}.

Similar to previous work, we also consider the scenario of image captured under collocated light and reconstruct a surface mesh as well as a SVBRDF.
However, we particularly focus on adaptively transferring details using inverse rendering only into regions that require a higher resolution to keep the overall representation more compact.

\section{Preliminaries}
\label{sec:preliminaries}

\begin{figure*}[t]
    \centering
    \includegraphics[width=\linewidth]{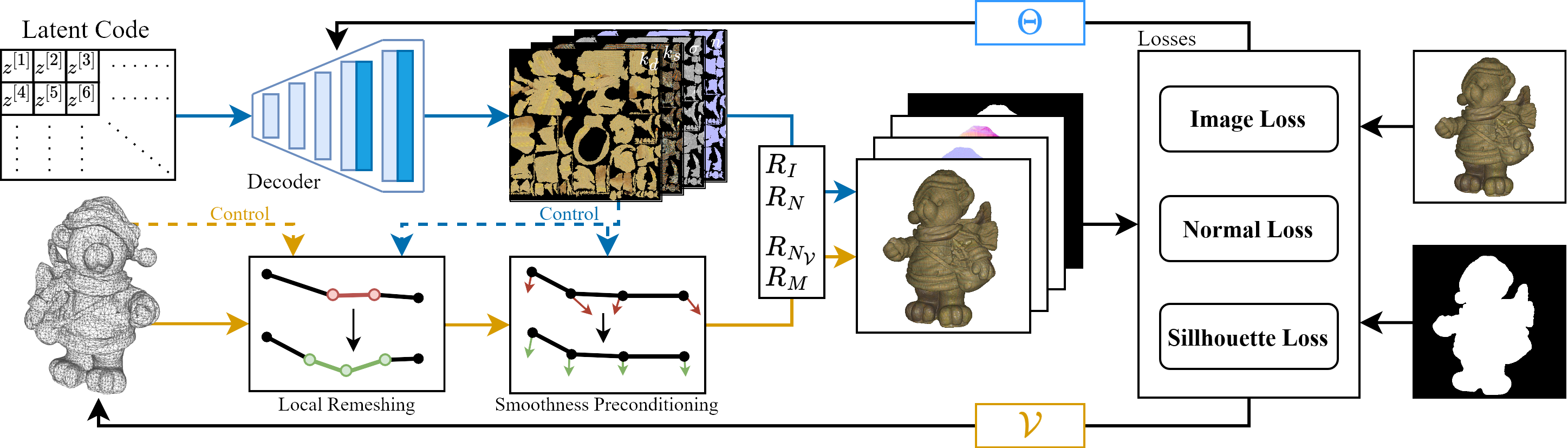}
    \caption{Overview of our inverse rendering framework, which performs decoder-based texture estimation and triangle mesh optimization with non-uniform smoothing. 
    The regional increase of the mesh resolution and its smoothness during optimization are controlled by the estimated texture normal and the current mesh curvature.}
    \label{fig:overview}
\end{figure*}

Given a set of images $\mat{I}$ captured under collocated light as well as respective masks $\mat{M}$ and camera parameters $\mathcal{C}$, we aim to reconstruct the shape and appearance of the observed object.
We represent the shape as a triangle mesh ${\mathcal{M} = (\mathcal{V}, \mathcal{F}, P)}$ consisting of a set of vertices ${\mathcal{V} = \{ \vec{v}_i \in \mathbb{R}^3 \}}$, a set of faces $\mathcal{F}$, as well as a UV-parameterization ${P \colon \mathcal{V} \rightarrow [0, 1]^2}$ for mapping the textures onto it.
For the material model, we use the Cook-Torrance model microfacet BRDF with a GGX normal distribution~\cite{walter2007microfacet} and store the corresponding parameters into a texture atlas ${(\vec{\kappa}_d, \vec{\kappa}_s, \sigma, \vec{n}) \colon [0,1]^2 \rightarrow \mathbb{R}^{10}}$, \ie diffuse albedo, specular albedo, roughness, and normals.

In our inverse rendering method of mesh and appearance, we use the nvdiffrast~\cite{laine2020modular} as a foundation for differentiable rasterization.
For this type of procedure, a suitable initialization of the mesh is important, so we obtain an initial estimate by computing the visual hull based on the input masks where its resolution is determined by a user-selected parameter.
To reconstruct the appearance, a fixed-size texture atlas is generated whose resolution is determined based on the input images by computing the footprint of the pixels in the UV space and projecting them onto the initial mesh.
In the inverse rendering context, the texture atlas is typically estimated by generative networks or auto-encoders, which require a reasonable initialization and are highly regularized by their training data and limited in terms of a fixed resolution by their architecture.

\section{Method}
\label{sec:method}

An outline of our method is depicted in \cref{fig:overview}.
We estimate the appearance features by optimizing the weights $\mat{\Theta}$ of a neural network and by generating textures of varying resolutions via texture tiles, which is further described in \cref{sec:appearance-texture-reconstruction}.
Furthermore, we optimize the vertex positions $\set{V}$ and adaptively refine the mesh resolution while ensuring smoothness, as shown in \cref{sec:mesh-geometry-reconstruction}.
This adaptive control is determined by a curvature-based criterion, explained in \cref{sec:controlling-the-surface-reconstruction}, to ensure that the mesh stays compact while details can still be transferred to regions where a higher resolution is required.
We infer all unknown parameters jointly in an inverse rendering framework based on a common image and silhouette losses which are complemented by a novel normal loss, defined in \cref{sec:objectives}, that bridges the detail transfer.

\subsection{Appearance Texture Reconstruction}
\label{sec:appearance-texture-reconstruction}

\begin{figure}[t]
    \centering
    \includegraphics[width=\linewidth, keepaspectratio]{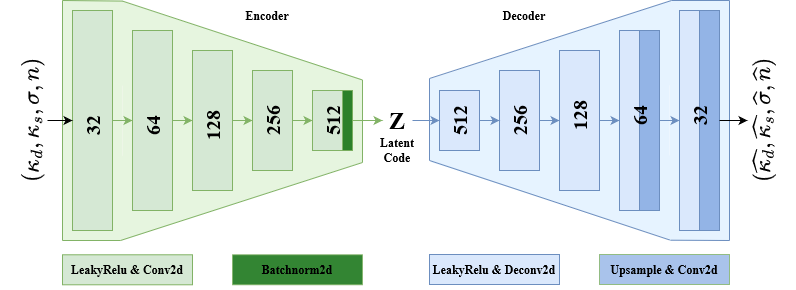}
    \caption{Architecture of our decoder for texture estimation which is pre-trained as part of an autoencoder and later finetuned. }
    \label{fig:decoder_architecture}
\end{figure}

To generate a texture atlas, we train a decoder $D_{\mat{\Theta}}$ during optimization, whereby a training from scratch could lead to potentially implausible SVBRDFs and, thus, suboptimal reconstructions.
Therefore, we pre-train $D_{\mat{\Theta}}$ as part of an autoencoder on a large material database~\cite{kaltheuner2021capturing,guo2020materialgan} to encourage that even random latent codes correspond to plausible SVBRDFs.
We overcome the typical resolution limitation of the networks by, instead of estimating a single high-resolution texture, utilizing $K$ fixed-size texture tiles with randomly sampled latent codes $(\vec{z}^{[1]}, \dots, \vec{z}^{[K]})$:
\begin{equation}
     (\vec{\kappa}_d^{[k]}, \vec{\kappa}_s^{[k]}, \sigma^{[k]}, \vec{n}^{[k]}) = D_{\mat{\Theta}}(\vec{z}^{[k]})
\end{equation}
Afterwards, we combine the individual tiles into a single high resolution texture, which is later used for rendering.
We prevent seams and other visible artifacts by using a small overlap between the tiles and blend these regions using a sigmoid weighting function.
All tiles have a resolution of $ { 128 \times 128 } $ and are inferred by the same decoder, whereby the entire encoder-decoder architecture is depicted in \cref{fig:decoder_architecture}.
Here, we use 5 convolution layers with LeakyReLU activations for the encoder and the respective deconvolution counterparts for the decoder.
In order to obtain textures which are highly detailed yet still smooth, we add an additional layer after the two last convolutions of the decoder which consists of an upsampling that is immediately followed by a downsampling convolution to propagate information from neighboring pixels.

\subsection{Mesh Geometry Reconstruction}
\label{sec:mesh-geometry-reconstruction}
\begin{figure}[t]
    \centering
    \includegraphics[width=\linewidth, keepaspectratio]{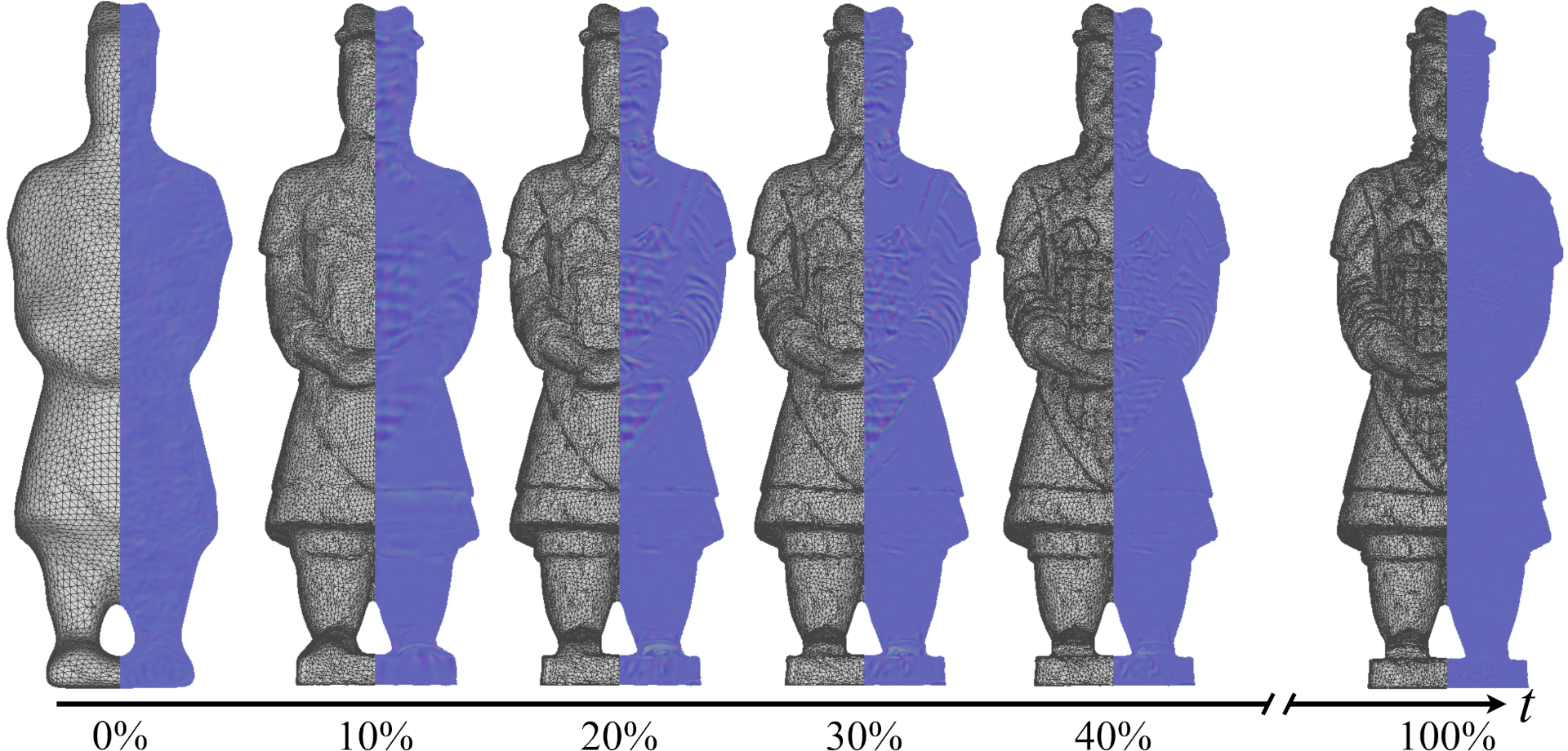}
    \caption{Geometry reconstruction over time. While in the early phase surface details are represented in the texture normals, these details are progressively transferred to the geometry in the later phase when the overall geometry is converging to the true shape.}
    \label{fig:geometry_timeline}
\end{figure}

Similar to the appearance estimation, the reconstruction of the surface geometry solely based on image data is challenging as reconstruction artifacts such as self-intersections may be introduced during optimization.
To avoid such issues, we first produce suitable reflection details via texture normals and combine the geometry optimization with a dedicated smoothness regularization.
Furthermore, we locally increase the mesh resolution and reduce the smoothness regularization to transfer the detailed reflection behavior to the geometry, as shown in \cref{fig:geometry_timeline}.

\paragraph*{Increasing Geometry Resolution.}
We start with a coarse mesh and aim to refine it in areas lacking geometry detail. 
These areas are identified in regions where the appearance reconstruction requires strongly varying normal maps for accurate representation.
Thus, we increase the mesh resolution as part of the inverse rendering process, considering both the current geometry and the normal map.
Specifically, we identify and adjust edges classified as too long based on individual edge lengths, see \cref{sec:controlling-the-surface-reconstruction}, during reconstruction. 
Afterwards, the edges are subdivided using LOOP~\cite{loop1987smooth} and the UV coordinates of the new vertex are calculated by interpolating the UV coordinates of the vertices that were previously connected.
By applying subdivision, we do not only insert a new vertex but also smooth the positions of the surrounding which, in turn, temporarily decreases the curvature in this region and prevents too excessive increases in resolution.
This is important as we only increase the resolution of the mesh and not reduce it later.
Additionally, we keep the number of edges per vertex approximately uniform by performing further edge flips to balance the edge connections in its neighbourhood.

\paragraph*{Local Surface Smoothness.}
Ensuring surface smoothness is a key objective during reconstruction and crucial to avoid self-intersections or similar artifacts on the surface.
Previous approaches typically aim to reconstruct meshes with uniform resolution~\cite{nicolet2021large}, \eg by efficiently smoothing the vertex movement across their neighbourhoods during optimization, which is encoded in the Laplacian matrix $\mat{L}$, based on a smoothness parameter $\lambda$ and a learning rate $\tau$:
\begin{equation}
    \vec{v} \leftarrow \vec{v} - \tau \, (\mat{\identity} + \lambda \, \mat{L})^{-2} \, \frac{\partial\mathcal{L}}{\partial \vec{v}}
\end{equation}
Our goal is to maintain smoothness also for non-uniform meshes where planar regions should be constrained more than highly curved regions.
We therefore do not rely on a single $\lambda$ value but incorporate adaptive values $\lambda_i(t)$ for each vertex which depend on the current normalized epoch ${t \in [0, 1]}$ to account for the continuous evolution of the surface. 
With this, we optimize the vertex positions using a locally preconditioned update:
\begin{gather}
    \vec{v} \leftarrow \vec{v} - \tau \, (\mat{\identity} + \mat{\Lambda}_{\mathcal{V}} \, \mat{L})^{-2} \, \frac{\partial\mathcal{L}}{\partial \vec{v}}
    \\
    \mat{\Lambda}_{\mathcal{V}}
    =
    \diag\left(\lambda_1(t), \dots, \lambda_{\abs{\mathcal{V}}}(t)\right)
    \in \mathbb{R}^{\abs{\mathcal{V}} \times \abs{\mathcal{V}}}
    \label{eq:lambda_matrix}
\end{gather}
Note that, in contrast to the original preconditioning, the resulting matrix $(\mat{\identity} + \mat{\Lambda}_{\mathcal{V}} \, \mat{L})$ is no longer symmetric and positive definite such that the above equation system needs to be solved with, \eg, a BICGSTAB solver.

\subsection{Controlling the Surface Reconstruction}
\label{sec:controlling-the-surface-reconstruction}
During optimization, we control the surface reconstruction by adjusting the smoothing variable $\lambda_i(t)$ and the mesh resolution via the edge measure $e_i(t)$ for each vertex $\vec{v}_i$.
To perform the local adjustments, we require clues about the surface details which we determine by considering the mesh curvature and a curvature derived from the texture normals.
Especially in early phases of the optimization when the surface is still coarse and kept smooth, it is important to rely on the texture normals which capture geometric details.
We estimate this texture space curvature $c_{\mat{n}}(\vec{v}_i)$ by applying a convolution with a Scharr kernel in $u$ and $v$ direction:
\begin{equation}
    c_{\mat{n}}(\vec{v}_i) = \norm{\frac{1}{2} [K_u * \vec{n} + K_v * \vec{n}](P(\vec{v}_i))}_2
\end{equation}
The mesh curvature $c_{\mathcal{V}}(\vec{v}_i)$ is computed using the Laplacian matrix $\mat{L}$:
\begin{equation}
    c_{\mathcal{V}}(\vec{v}_i) = \frac{1}{2} \norm{(\mat{L} \, \vec{v})_i}_2 
\end{equation}

\paragraph*{Reconstruction Uncertainties.}
As we continuously update the mesh based on intermediate reconstruction results, we have to make sure that the clues we infer from the surface are not a result of an uncertain or far from optimal state, especially of states stemming from the initialization.
We therefore start the optimization by first focusing on the silhouette loss to improve the initial coarse geometry based on the input masks.
Over the course of the optimization, the weight of the image loss and normal loss are damped and we progressively increase them over time.
This loss damping is implemented by additional weights for the image and normal loss, based on a scaled sigmoid function to ensure a smooth transition which is controlled by the normalized epoch ${t \in [0, 1]}$ of the optimization.
\begin{equation}
   s_{\mathrm{loss}}(t) = \mathrm{sigmoid}(20 (t\!-\!0.2)) = \frac{1}{1 + \euler^{-20 (t - 0.2)}}
\end{equation}
To further increase the stability, we additionally delay the increase of geometry resolution and the decrease of smoothness in the optimization process to first capture as much detail as possible within the coarser geometry.
\begin{equation}
   s_{\mathrm{control}}(t) = \mathrm{sigmoid}(20 (t - 0.3)) = \frac{1}{1 + \euler^{-20 (t - 0.3)}}
\end{equation}
Note that this control damping is further shifted in time in comparison to the loss damping to encourage that the resolution is increased only after details have been transferred.
Furthermore, to handle uncertain intermediate optimization steps, we do not perform the local remeshing steps in every epoch but rather use every 25 epochs.
Note that we still compute the curvatures $c_{\mat{n}}(\vec{v}_i)$ and $c_{\mathcal{V}}(\vec{v}_i)$ in each epoch and additionally average them over this window to get a more robust quantity for mesh adaption.

\begin{figure}[t]
    \centering
    \includegraphics[width=0.8\linewidth, keepaspectratio]{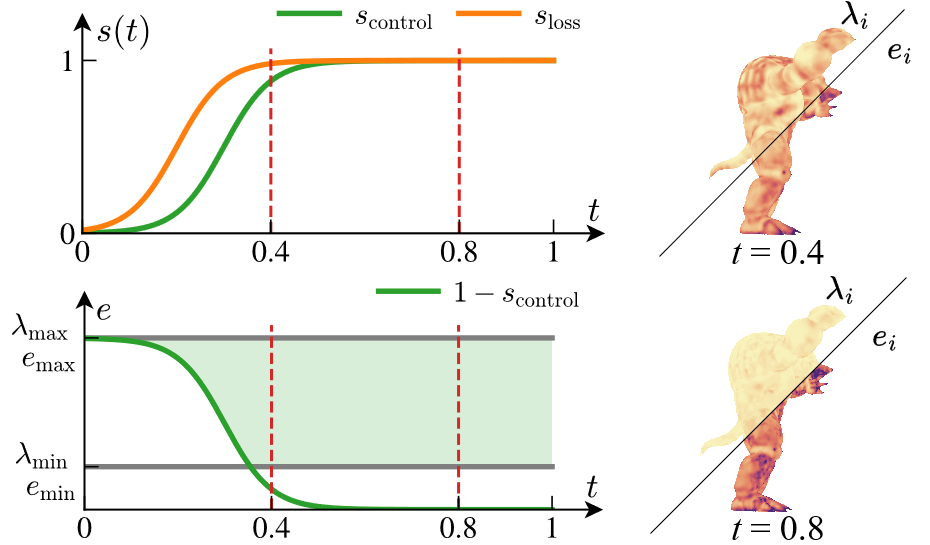}
    \caption{Damping functions used to increase the stability of the reconstruction and to transfer the geometry information from the texture normal to the mesh surface.}
    \label{fig:epoch_scaling}
\end{figure}

\paragraph*{Regional Smoothness Adjustments.}
We base the adjustment of the smoothness preconditioning on our texture space curvature $c_{\vec{n}}(\vec{v}_i)$ to lesser condition the vertex movement as long as larger updates are needed.
Thus, we consider the inverted curvature
\begin{equation}
    s_{\lambda}(\vec{v}_i) = 1 - \clip( w_{\mat{n}} \cdot c_{\mat{n}}(\vec{v}_i) ) \in [0, 1]  
    \label{eq:lambda_scaling}
\end{equation}
as a measure to control this process, where $ { w_{\mat{n}} = 3 } $ and $\clip$ is the operator to clamp the values into the range $ [0, 1] $.
We define the per-vertex $\lambda_i(t)$ value, which is required to compute $\mat{\Lambda}_{\mathcal{V}}$ in \cref{eq:lambda_matrix}, based on $ s_{\lambda}(\vec{v}_i) $:
\begin{equation}
    \lambda_i(t) = \lerp(\lambda_{\mathrm{min}}(t), \lambda_{\mathrm{max}}(t), s_{\lambda}(\vec{v}_i))
    \label{eq:lambda_computation}
\end{equation}
Here, $ \lerp $ is the standard linear interpolation operator.
We require a time-dependent lower bound $ { \lambda_{\mathrm{min}}(t) = \max\{ \lambda_{\mathrm{min}}, (1 - s_{\mathrm{control}}(t)) \cdot \lambda_{\mathrm{max}} \} }$ and a constant upper bound $ { \lambda_{\mathrm{max}}(t) = \lambda_{\mathrm{max}} } $ to enforce a stronger smoothness in the early phases of the reconstruction and gradually relax the enforcement, based on $s_{\mathrm{control}}(t)$, over time, visualized in \cref{fig:epoch_scaling}.
The parameters $\lambda_{\mathrm{min}}$ and $\lambda_{\mathrm{max}}$ are set to $16$ and $64$ respectively.

\paragraph*{Regional Mesh Resolution.}
Similarly, we aim to locally control the mesh resolution to adapt when texture information is transferred into the geometry, but this process should not be solely based on $ c_{\mat{n}}(\vec{v}_i) $, but also include the current curvature of the geometry $ c_{\mathcal{V}}(\vec{v}_i) $:
\begin{equation}
    s_{e}(\vec{v}_i) = 1 - \clip( w_{\mathcal{V}} \cdot c_{\mathcal{V}}(\vec{v}_i) + w_{\mat{n}} \cdot c_{\mat{n}}(\vec{v}_i) ) \in [0, 1]  
\end{equation}
Here, $ w_{\mat{n}} $ is defined as above and $ { w_{\mathcal{V}} = \frac{1}{16} } $.
In the same way as \cref{eq:lambda_computation}, we define an edge length indicator function
\begin{equation}
    e_i(t) = \lerp(e_{\mathrm{min}}(t), e_{\mathrm{max}}(t), s_{e}(\vec{v}_i))
\end{equation}
using a similarly defined time-dependent lower bound $ { e_{\mathrm{min}}(t) = \max\{ e_{\mathrm{min}}, (1 - s_{\mathrm{control}}(t)) \cdot e_{\mathrm{max}} \} } $ and a constant upper bound $ { e_{\mathrm{max}}(t) = e_{\mathrm{max}} } $, visualized in \cref{fig:epoch_scaling}.
The values $e_i$ are smoothed over their neighborhoods to produce a smoother local edge length indicator.
They are then used to decide whether a surrounding edge of a vertex $\vec{v}_i$ should be split by averaging the per vertex indicator values along an edge to compute per edge values.
The split is performed according to the edge split method, see \cref{sec:mesh-geometry-reconstruction}, if the actual length of an edge is longer than $e_v$.
Furthermore, we combine this splitting with edge flipping~\cite{Palfinger2022} to keep the ideal number of associated neighbours per vertex~\cite{botsch04}.
We scale the initial mesh to a normalized space in range $[-1,1]$, which enables us to determine a minimum and maximum edge length $ { e_{\mathrm{min}} = 0.01875} $ and $ { e_{\mathrm{max}} = 0.375} $.

\begin{figure}[t]
    \centering
    \includegraphics[width=\linewidth]{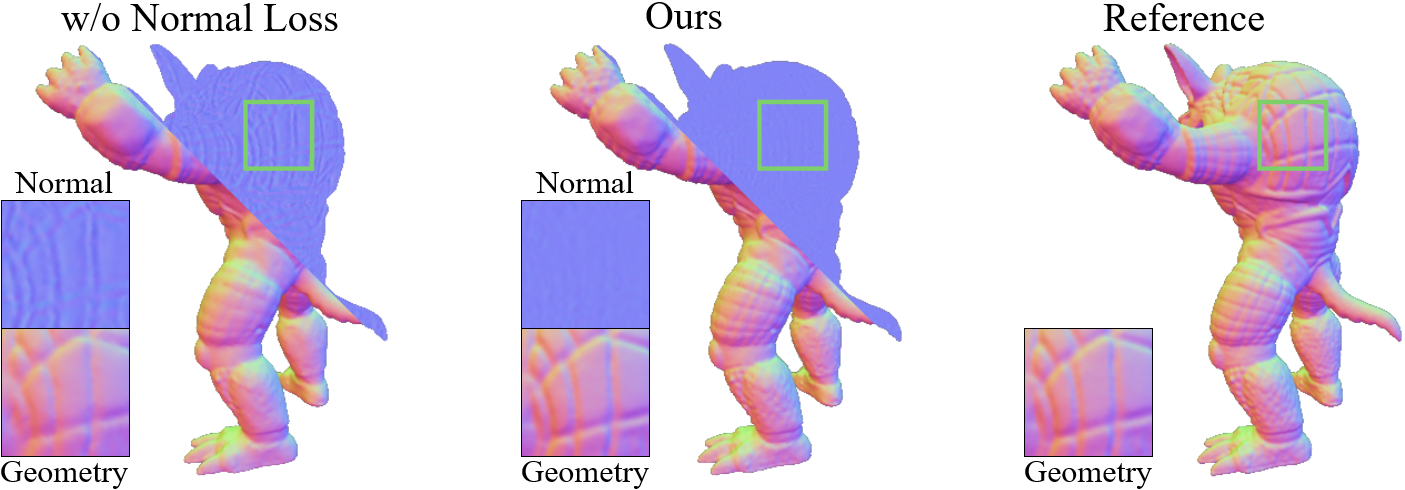}
    \caption{Our normal loss supports the transfer of information from the texture normal to the geometry.}
    \label{fig:losses}
\end{figure}

\subsection{Objectives}
\label{sec:objectives}

The overall optimization is based on three losses:
\begin{equation}
    \mathcal{L} = w_{\mathrm{img}} \, \mathcal{L}_{\mathrm{img}} + w_{\mathrm{sil}} \, \mathcal{L}_{\mathrm{sil}} + w_{\mathrm{normal}} \, \mathcal{L}_{\mathrm{normal}}
\end{equation}
The image loss computes the masked $L_1$ difference between the input images $\mat{I}$ and the rendered images $\widehat{\mat{I}}$ and $\widehat{\mat{I}}_{\mathcal{\vec{V}}}$:
\begin{align}
    \mathcal{L}_{\mathrm{img}} = \ & \frac{1}{2} \norm{\mat{m} \odot \left(\log(\widehat{\mat{I}} + 0.01) - \log(\mat{I} + 0.01)\right)}_1 \nonumber \\
    + \ & \frac{1}{2} \norm{\mat{m} \odot \left(\log(\widehat{\mat{I}}_{\mathcal{\vec{V}}} + 0.01) - \log(\mat{I} + 0.01)\right)}_1
\end{align}
Meanwhile, the silhouette loss computes the $L_1$ difference between the input masks $\mat{M}$ and the estimated masks $\widehat{\mat{M}}$:
\begin{equation}
    \mathcal{L}_{\mathrm{sil}} = \norm{\widehat{\mat{M}} - \mat{M}}_1
\end{equation}
Our novel normal loss, visualized in \cref{fig:losses}, is not defined on any of the inputs, as it penalizes the masked $L_1$ difference between the rendered surface normal $\widehat{\mat{N}}_{\mathcal{\vec{V}}}$ and rendered perturbed surface normal ${ \sg[\widehat{\mat{N}}_{\mathcal{\vec{V}}}] + \widehat{\mat{N}} } $:
\begin{equation}
    \mathcal{L}_{\mathrm{normal}} = \norm{\mat{m} \odot \left((\sg[\widehat{\mat{N}}_{\mathcal{\vec{V}}}] + \widehat{\mat{N}}) - \widehat{\mat{N}}_{\mathcal{\vec{V}}}\right)}_1
\end{equation}
Here, $ \sg $ is the stop-gradient operator which returns the argument unaltered but sets all partial derivatives to zero.
This avoids that the term $\widehat{\mat{N}}_{\mathcal{\vec{V}}}$ cancels out in the loss.
It is also used to calculate masks without silhouette gradients:
\begin{equation}
    \mat{m} = \mat{M} \odot \sg[\widehat{\mat{M}}]
\end{equation}

We generate the necessary images using in the above losses by the differentiable rendering operator $R$:
\begin{gather}
    \widehat{\mat{I}} = R_{\mat{I}}(\mathcal{V}, \mat{\Theta} \given \mathcal{F}, P, \vec{z})
    , \quad
    \widehat{\mat{I}}_{\mathcal{\vec{V}}} = R_{\mat{I}}(\mathcal{V}, \mat{\theta} \given \mathcal{F}, P, \vec{z}),\\
    \widehat{\mat{M}} = R_{\mat{M}}(\mathcal{V} \given \mathcal{F}), \\
    \widehat{\mat{N}} = R_{\mat{N}}(\mat{\Theta} \given \mathcal{V}, \mathcal{F}, P, \vec{z})
    , \quad
    \widehat{\mat{N}}_{\mathcal{\vec{V}}} = R_{\mat{N}_{\mathcal{\vec{V}}}}(\mathcal{V} \given \mathcal{F})
\end{gather}
While $\widehat{\mat{I}}$ is applied to optimize the mesh vertices $\mathcal{V}$ as well as the decoder weights $\mat{\Theta}$, incorporating $\widehat{\mat{I}}_{\mathcal{\vec{V}}}$ aims to reduce the impact of the texture normals by not using them during rendering and, thus, effectively only optimizing the subset $ { \mat{\theta} \subset \mat{\Theta} } $ of the decoder weights, which is used for this rendering.
In addition to propagating gradients from the rendering of the surface normals $R_{\mat{N}_{\mathcal{\vec{V}}}}$ to the vertex positions $\mathcal{V}$, we also propagate gradients to the decoder weights $\mat{\Theta}$ via the texture normal rendering $R_{\mat{N}}$.

Since generating a coarse geometry reconstruction is necessary, we weight the silhouette loss $ { w_{\mathrm{sil}} = 1 } $, while the image and normal loss terms are weighted by ${ w_{\mathrm{img}} = 0.05 } $ and ${ w_{\mathrm{normal}} = 0.01 } $ respectively, for synthetic data and lower these further to $ \num{1e-3} $ and $ \num{1e-4} $ for real world data, where the respective losses are typically higher, due to limitations of our material model.

\section{Evaluation}
\label{sec:label}
\begin{figure}[t]
    \centering
    \includegraphics[width=\linewidth]{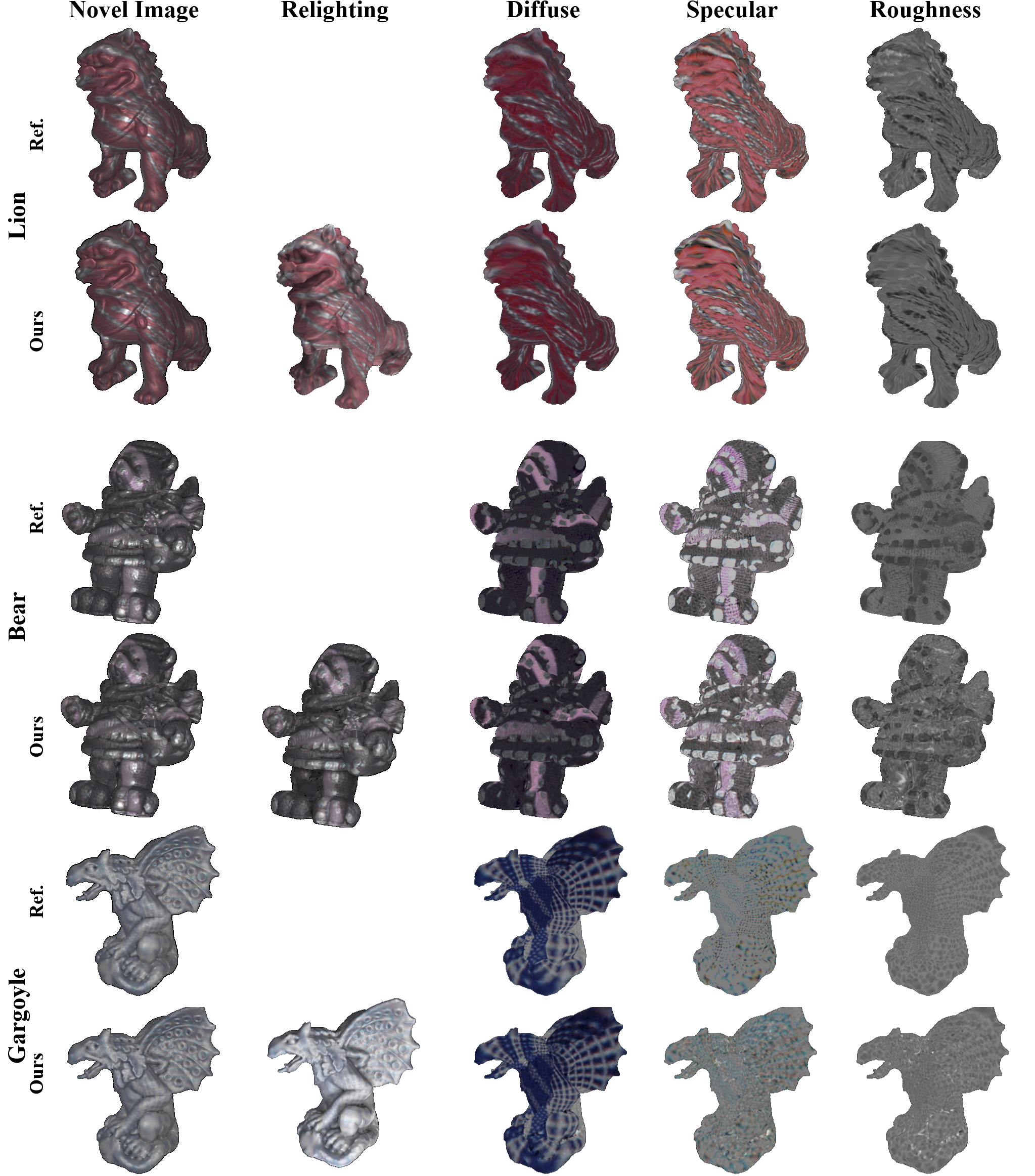}
    \caption{Reconstructions of synthetic input data based on 50-70 input images with a resolution of $512^2$. Please zoom in for details.}
    \label{fig:synthetic}
\end{figure}

We compare our adaptive reconstruction method, based on synthetic and real world collocated capture scenarios.
Therefore, we create synthetically generated test and train data using publicly available 3D objects as well as the validation split of our material dataset, resulting in a total of 14 synthetic input objects.
To also perform a quantitative analysis, we evaluate our estimation in \cref{tab:want} based on rendered features mapped onto the surface to account for the difference in mesh parameterization which is measured by the MSE, SSIM~\cite{wang2004ssim}, LPIPS~\cite{zhang2018lpips}, and PSNR.
Additionally, we also evaluate the quality of the reconstructed geometry based on the Chamfer distance (CD) and Hausdorff distance (HD). 
In \cref{fig:synthetic}, we show the comparison of reference test data with our reconstruction and show additional relightings in a supplemental video.
We are able to produce highly detailed results, with mesh resolutions ranging from $30$k to $71$k vertices for more complex objects, requiring $\sim$\,1\,-\,4\,h for 50 to 170 images on a NVIDIA A100 GPU.
\begin{figure}[t]
    \centering
    \includegraphics[width=\linewidth]{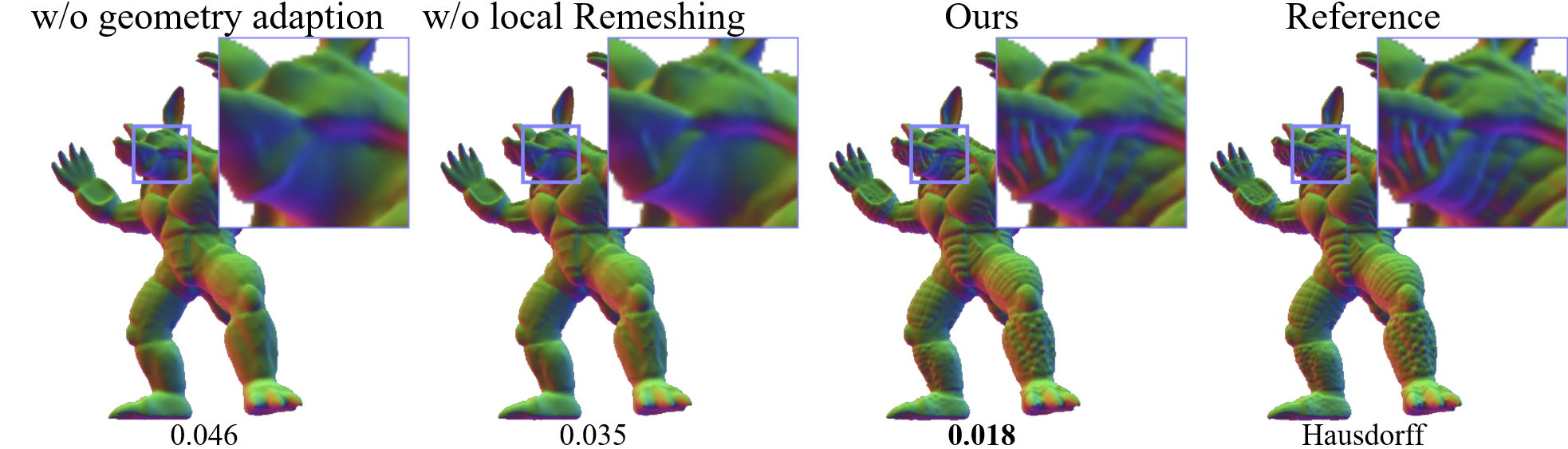}
    \caption{Our adaptive geometry reconstruction approach, with the local remeshing or smoothness adaption disabled. Our approach, combining both adaption methods, produces most detailed results.}
    \label{fig:geometry}
\end{figure}

\begin{table}
    \scriptsize
    \centering
    \caption{Quantitative evaluation of the synthetic test dataset on test images as well as reference appearance features.}
    \renewcommand{\arraystretch}{1.075}
    \begin{tabular}{l|cccc|cc}
        \toprule
         & \multicolumn{4}{c|}{PSNR $\uparrow$} & CD $\downarrow$ & HD $\downarrow$ \\
         & Image & $\boldsymbol{\kappa}_d$ & $\boldsymbol{\kappa}_s$ & $\sigma$ & & \\
        \midrule
        Lion & 40.51 & 43.95 & 33.53 & 33.19 & 0.015 & 0.032 \\
        Bear & 37.77 & 40.87 & 31.20 & 30.05 & 0.014 & 0.159 \\
        Gargoyle & 42.29 & 37.26 & 33.18 & 35.95 & 0.017 & 0.161 \\
        \midrule
        Dataset & 41.40 & 42.72 & 34.55 & 34.90 & 0.027 & 0.120 \\
        \bottomrule
    \end{tabular}
    \label{tab:want}
\end{table}

\begin{figure*}[t]
    \centering
    \includegraphics[width=\linewidth]{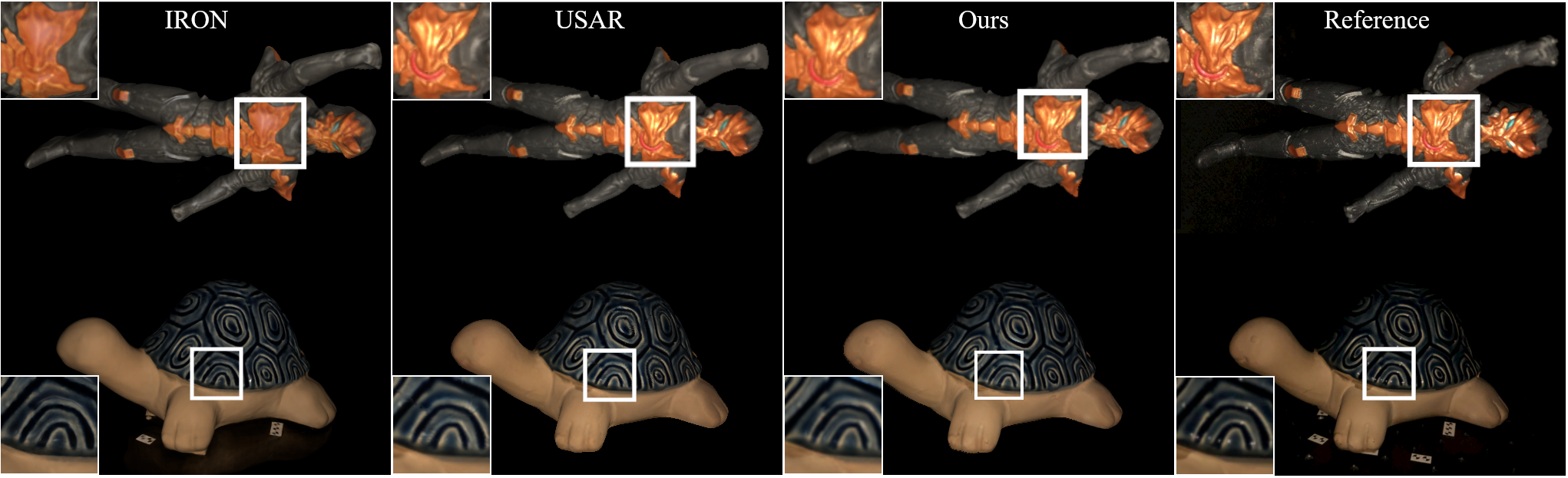}
    \caption{Comparison of reconstruction results on real-world objects for collocated capture scenarios based on 100 (superman) and 170 (turtle) images respectively recorded with a smartphone.}
    \label{fig:realworld}
\end{figure*}

\subsection{Real World Evaluation}
We evaluate our method based on a real world dataset and compare it to the results of two other reconstruction methods~\cite{kaltheuner2023unified}~\cite{zhang2022iron}, which both reconstruct meshes with uniform resolution, fixed texture resolution and make use of collocated capture scenarios.
The real world dataset by Luan \etal~\cite{luan2021unified} contains data captured with a smartphone camera and flashlight.
The results, presented in \cref{fig:realworld}, show great results for the superman figurine compared to both other methods while using fewer vertices than USAR~\cite{kaltheuner2023unified} and less than a tenth of the mesh resolution of IRON~\cite{zhang2022iron}.
Our method and USAR are able to reproduce the highlight almost perfectly, meanwhile IRON is not able to produce the highlight at all.
Our turtle figure reconstruction also requires fewer vertices, as shown in \cref{tab:real}, but outperforms the other methods in terms of MSE and LPIPS values.
While IRON is able to reconstruct the object, it also reconstructs the floor as it does not require any input masks.

\begin{table}
    \scriptsize
    \centering
    \caption{Quantitative evaluation of real world data reconstructions averaged on the 53 (superman) and 43 (turtle) test images.}
    \renewcommand{\arraystretch}{1.25}
    \begin{tabular}{l|l|cccc|c}
        \toprule
        \multicolumn{1}{c}{} & & MSE $\downarrow$ & SSIM $\uparrow$ & LPIPS $\downarrow$ & PSNR $\uparrow$ & $\abs{\mathcal{V}}$ $\downarrow$ \\
        \midrule
        \multirow{3}{*}{\rotatebox{90}{superman}}
        & IRON~\cite{zhang2022iron}& 0.0015 & 0.942 & 0.065 & 28.70 & 1.206k \\
        & USAR~\cite{kaltheuner2023unified} & \textbf{0.0008} & 0.954 & 0.054 & \textbf{31.40} & \phantom{0.}139k \\
        & Ours & \textbf{0.0008} & \textbf{0.959} & \textbf{0.036} & 30.93 & \phantom{0.0}\textbf{28k} \\
        \midrule
        \multirow{3}{*}{\rotatebox{90}{turtle}}
        & IRON~\cite{zhang2022iron} & 0.0036 & \textbf{0.921} & 0.118 & 25.51 & 1.107k \\
        & USAR~\cite{kaltheuner2023unified} & 0.0034 & 0.918 & 0.125 & \textbf{26.17} & \phantom{0.}139k \\
        & Ours & \textbf{0.0031} & 0.919 & \textbf{0.085} & 25.09 & \phantom{0.0}\textbf{56k} \\
        \bottomrule
    \end{tabular}
    \label{tab:real}
\end{table}
\subsection{Ablation Studies}
\label{sec:ablation_study}

In our work, we present a method which focuses on the reconstruction of highly detailed objects with small geometry and texture resolution.
Our texture estimation approach is flexible in regards to resolution, as it operates in a tile-based fashion and due to the ability of the decoder to output textures of different resolutions controlled by the resolution of the latent variable $\vec{z}$, see \cref{sec:appearance-texture-reconstruction}.
However, the choice of arbitrarily high output texture resolutions has drawbacks, as the input resolution limits the texture resolution.
Respective results are shown in the supplemental material.

To perform an ablation study on our geometry reconstruction, we discuss the impact of our mesh adaption methods.
We introduce our per vertex smoothness to increase the level of detail in specific regions, while keeping other regions smooth which comes with a change of the underlying solver from Cholesky to BIGCGSTAB.
Additionally, we introduce our coarse-to-fine optimization by local remeshing of regions which require more detail.
Turning off the local remeshing, as depicted in \cref{fig:geometry}, results in an less detailed reconstruction, where our local smoothness adjustment is able to introduce some details.
Meanwhile, removing both of our geometry altering methods results in extremely low detailed geometry reconstruction.
Our locally adaptive geometry optimization approach is able to produce highly detailed geometry reconstructions.
We have also tested the behavior of our method with different damping functions in the supplementary material.

\begin{figure}[t]
    \centering
    \includegraphics[width=\linewidth]{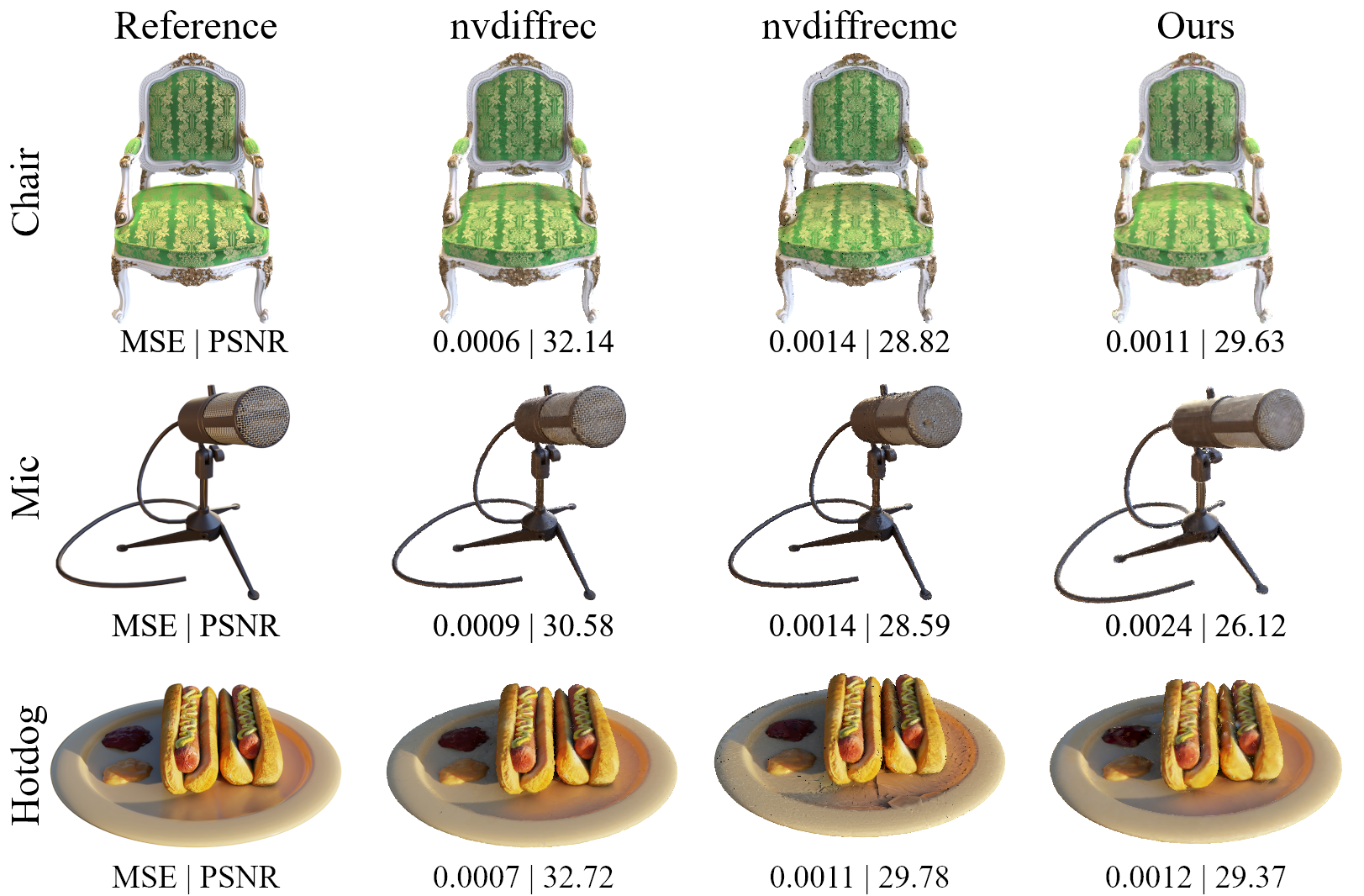}
    \caption{Comparison of reconstruction results on NeRF datasets (100 images each) where the environment light has been estimated as well. Even when replacing the rasterizer in our pipeline with a differentiable Monte-Carlo renderer and using a different material model that has not been used in the pre-training of our appearance decoder network, our approach achieves similar results.}
    \label{fig:nvdiffrec}
\end{figure}

So far, we have focused on collocated capture scenarios with a rasterization-based rendering method.
However, our approach is in principle not limited in this regard and can also be extended to less restrictive scenarios as well as different underlying rendering approaches. 
For this purpose, we integrated the differentiable Monte Carlo renderer of nvdiffrecmc~\cite{hasselgren2022shape} as well as the accompanying environment light estimation approach and conducted a further comparison against state-of-the-art methods~\cite{hasselgren2022shape,Munkberg_2022_CVPR} that do not require a predefined lighting scenario and are, thus, more suitable for capturing in the wild.
We evaluated on the popular NeRF dataset~\cite{mildenhall2020nerf} each comprising 100 input images with a resolution of ${400 \times 400}$ and show the results in \cref{fig:nvdiffrec}.

Our method achieves similar results as nvdiffrec~\cite{Munkberg_2022_CVPR} and nvdiffrecmc~\cite{hasselgren2022shape}, although it has not been particularly tuned for this illumination scenario.
In particular, we kept our pre-trained decoder unchanged and estimated materials for the different physically-based (PBR) model from Disney~\cite{mcauley2012practical}.
Here, we averaged the resulting $ { \vec{\kappa}_s \in [0, 1]^3 } $ to single-channel metallic values and relied on the decoder architecture to generate plausible results.
In line with the observations from nvdiffrecmc~\cite{hasselgren2022shape}, inferring reliable texture normals is challenging as the estimated environment illumination tends to bake the colors of the object material.
This could be mitigated with additional, intentionally left out regularization terms for the lighting and normals~\cite{hasselgren2022shape}.

\section{Conclusion}
\label{sec:conclusion}

We presented a novel inverse rendering method for the joint reconstruction of compact object shape with adaptive resolution and high resolution appearance textures.
For this, we obtained material properties in terms of a SVBBRDF along with normal maps by finetuning a single decoder network and employing a tiling-based generation of the texture atlas.
Furthermore, we transferred fine geometric details from the estimated normal maps onto the mesh geometry using a novel loss term and introduced a local refinement the mesh geometry using on a curvature-based criterion.
In the scope of a qualitative and quantitative evaluation, we demonstrated that our method is able to reconstruct more compact, high-quality models than previous methods.

\paragraph{Limitations and future work}
Our method is not without limitations, which need to be addressed in the future.
Other works have already explored the estimation of the light scenario alongside the reconstruction, which poses new challenges for the stability of our adaptive geometry reconstruction.
We do not model multiple light bounces, which can lead to estimation errors in the appearance reconstruction as well as the geometry reconstruction.

\section*{Acknowledgements}

This research has been funded by the Federal Ministry of
Education and Research under grant no. 01IS22094A
WEST-AI, by the Federal Ministry of Education and
Research of Germany as well as the state of North-Rhine
Westphalia as part of the Lamarr-Institute for Machine
Learning and Artificial Intelligence, INVIRTUO under grant no. PB22-063A and by
the DFG project KL 1142/11-2.

{\small
\bibliographystyle{ieee_fullname}
\bibliography{egbib}
}

\end{document}